# Jet noise characterization for advanced pipeline leak detection

Riccardo Angelo Giro[1] [*], Giancarlo Bernasconi[1], Simone Cesari[2], Giuseppe Giunta[2]
[1] Politecnico di Milano, Milan
[2] Eni S.p.A., San Donato Milanese
[*] Corresponding author (email: riccardoangelo.giro@polimi.it)

## ABSTRACT

The detection of leaks in pipeline transportation systems is a matter of serious concern for operators, who pursue the integrity of their assets, the reduction of losses and the prevention of environmental hazards. Whenever a hole occurs in a pressurized pipeline, the corresponding fluid leakage is characterized by a turbulent flow and a peculiar acoustic noise, whose characteristics depend also on the size of the hole itself. This study shows that both the presence and the size of such a leaking hole can be successfully detected, by exploiting the acoustic noise (pressure transients) generated by the fluid exiting the pipe and recorded internally by hydrophones, or by considering the corresponding vibrations (e.g., acceleration signals) propagating along the external shell of the conduit. To this purpose, several experimental campaigns of acoustic noise generation have been performed using multiple calibrated nozzles on a 16" ID connection pipeline in a fuel tanks area. Detection and classification procedures are proposed to control the presence of leakages and to estimate the size of the hole, using pressure and vibration signals.

## 1. INTRODUCTION

Jet noise is generated in the environment where the fluid jet is flowing. The possibility of recording this noise relies on the presence of an appropriate receiver (e.g. a microphone in air) close to the location of the hole, on the side of the exiting jet. A big theoretical and experimental research work has been done to compute such a noise produced by a fast fluid jet: the more important results are due to Lighthill [1, 2] for the acoustic emission due to a turbulent jet, and the spectral distribution of the acoustic emission. The work of Lighthill is focused on air jets, but the results are applicable also to other fluids, as long as turbulence remains the main source of noise. Experimental research studies have been carried out by Liang et al. [3], in the context of gas pipeline leakage detection, and by Kaewwaewnoi et al. [4] for internal valve leakage estimation. Both papers report Lighthill formulas and experimentally study the dependence of acoustic emission on operational parameters. Hutt [5] instead analyzed the noise produced by water jetting systems (used to clean or cut materials) and its dependence on the operational parameters. His conclusions are the following: high frequency energy (between 4 kHz and 8 kHz) dominates the water jet noise spectrum, and the sound pressure level (SPL) increases proportionally with water pressure and with the nozzle diameter. Moreover, the paper analyzes the jet noise characteristics at different recording angles with respect to the jet source.

The case study analyzed in this research differs from the works, since we employ vibroacoustic sensors in contact with the inner fluid and with the pipe shell. In practice, the acoustic noise we are trying to detect is produced by turbulence at the entering side of the hole. There is not a unique established literature on this topic, as the latter is typically modelled as a chaotic movement of particles flowing in a duct. One reference that validates the results with experimental data shows there is a parabolic relationship between the Sound Pressure Level (SPL) and the flow rate, as also shown by Xu et al. [6]. We also address the problem experimentally, simulating realistic scenarios of liquid/gas exiting from calibrated nozzles with different area and shape. We are expecting at least an increase of the acoustic noise for increasing differential pressure and hole cross area.

The following sections describe the experimental campaigns performed to measure the (acoustic) noise produced by a leaking hole. The leakage is generated by opening a branch where a calibrated nozzle is mounted, such that the exiting cross section and shape are known. The main objective is to verify the noise character versus the nozzle parameters, namely: the detectability of an existing leak by processing acoustic noise data; the localization of the leak position; the possibility of discriminating the shape/area of the hole.

## 2. CASE STUDY DESCRIPTION

This work makes use of pressure signals collected by a proprietary vibroacoustic monitoring system (e-vpms® technology [7, 8, 9]) installed on a fuel tanks deposit pipeline, property of Eni and located in North Italy. Such a line has a length of approximately 341 m and is characterized by 16" ID carbon steel pipes. A schematic representation of the conduit and the location of the recording stations are displayed in Fig. 1, respectively with a red line and yellow/blue circles. The six e-vpms® measurement units (labelled with the letters from A to F) are equipped with one or more of the following sensors:

- a pressure transducer, recording the absolute pressure of the transported fluid;
- a dynamic hydrophone, which measures small-scale dynamic pressure variations;
- an accelerometer, which detects the vibrations propagating along the pipe shell;
- a flowmeter, recording the instantaneous mass and volumetric flow rates. Table 1 reports the distance between each station with respect to one pipeline end (e.g., station A).

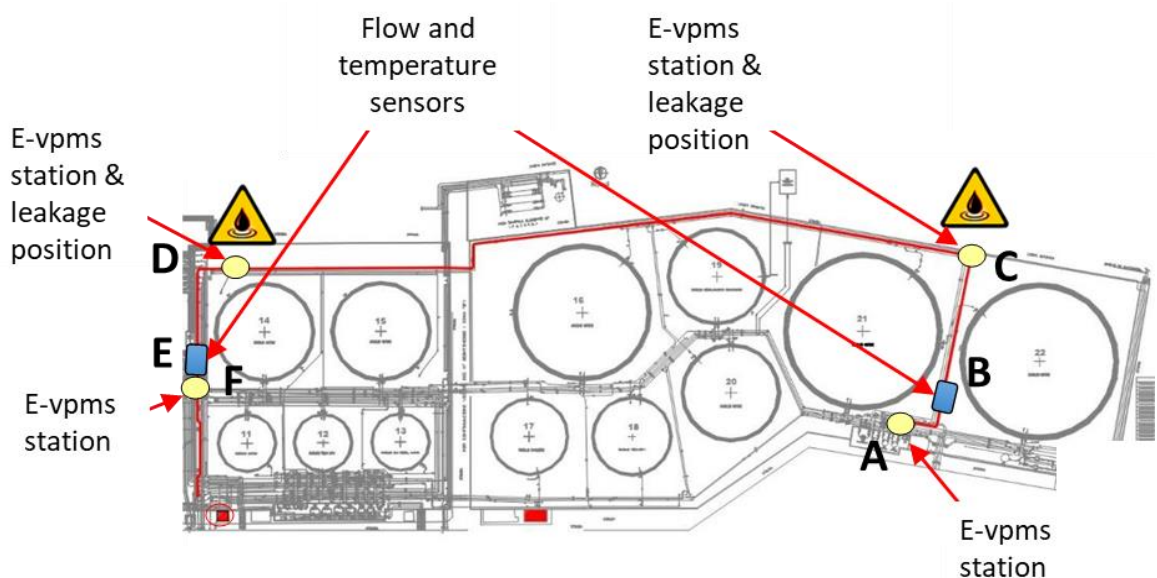

Fig. 1. Schematic of the pipeline track (red line) and of the e-vpms® stations (yellow and blue circles).

Table 1. Layout of the various stations distributed along the pipeline.

| Station | Distance with respect to station A (m) |
|---------|----------------------------------------|
| A | 0 |
| B | 10 |
| C | 63 |
| D | 294 |
| E | 337 |
| F | 341 |

The data were collected for more than a month. During the test campaigns, we have performed several controlled spills utilizing nozzles of varying shapes and sizes (as displayed in Fig. 2), employing different opening methods on two distinct locations along the pipeline: the corresponding spill valves have been respectively installed at stations C and D. Lastly, it should be noted that, in pairs, a circular and a slot hole have the same area.

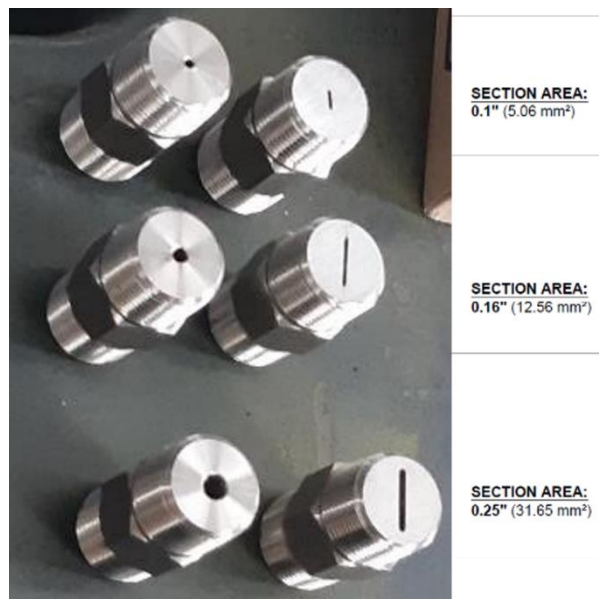

Fig. 2. Nozzles employed during the spill tests.

## 3. DATA ANALYSIS

By looking at the experiment data, we have identified that the fuel tanks deposit falls in two different operating conditions:

1. "Standstill", whenever the pumps are off, and the absolute pressure recorded by the static hydrophones is around 0.6 bar or less.
2. "Transferring", with pressures typically around 4 bar, yet having sporadic lows reaching 0.8 bar.

In general, these conditions can be detected by analyzing pressure sensors, hydrophones, and flowmeter data. The volumetric flow is zero when the pipeline is "standstill" and above 100 $m^3$/h in "transferring" condition. As an example, Fig. 3 shows the maximum value (max) and the standard deviation (std) of the measured pressure signals during the entire acquisition month, whereas Fig. 4 displays the flow rate at stations B and E within the same time frame.

At first sight, there is no visual evidence of the spill campaigns, yet it is easy to distinguish the day (high activity) from the night/holidays (low activity). We note here that the leakage tests occurred while the tank trucks were being filled with fluid: the latter events can be seen as high flow rate leakages from the monitored pipes to the tank on the truck, thus introducing a significant disturbance component in the experiments. As an example, Fig. 5 and Fig. 6 show a spill test performed at station C, as seen by the available pressure and flow rate signals. By looking at Fig. 5 and Fig. 6, we can make the following observations: there is a decrease of the static pressure levels at the leaking point; the acoustic noise produced by the spill is a high bandwidth signal, which extends up to 4 kHz; the spill noise is evident only in proximity of the recording site; even at around 60 m of distance from the source (e.g., station A), such a signal is not visible anymore, since the noise of the pumping systems is dominant and hides the leak itself.

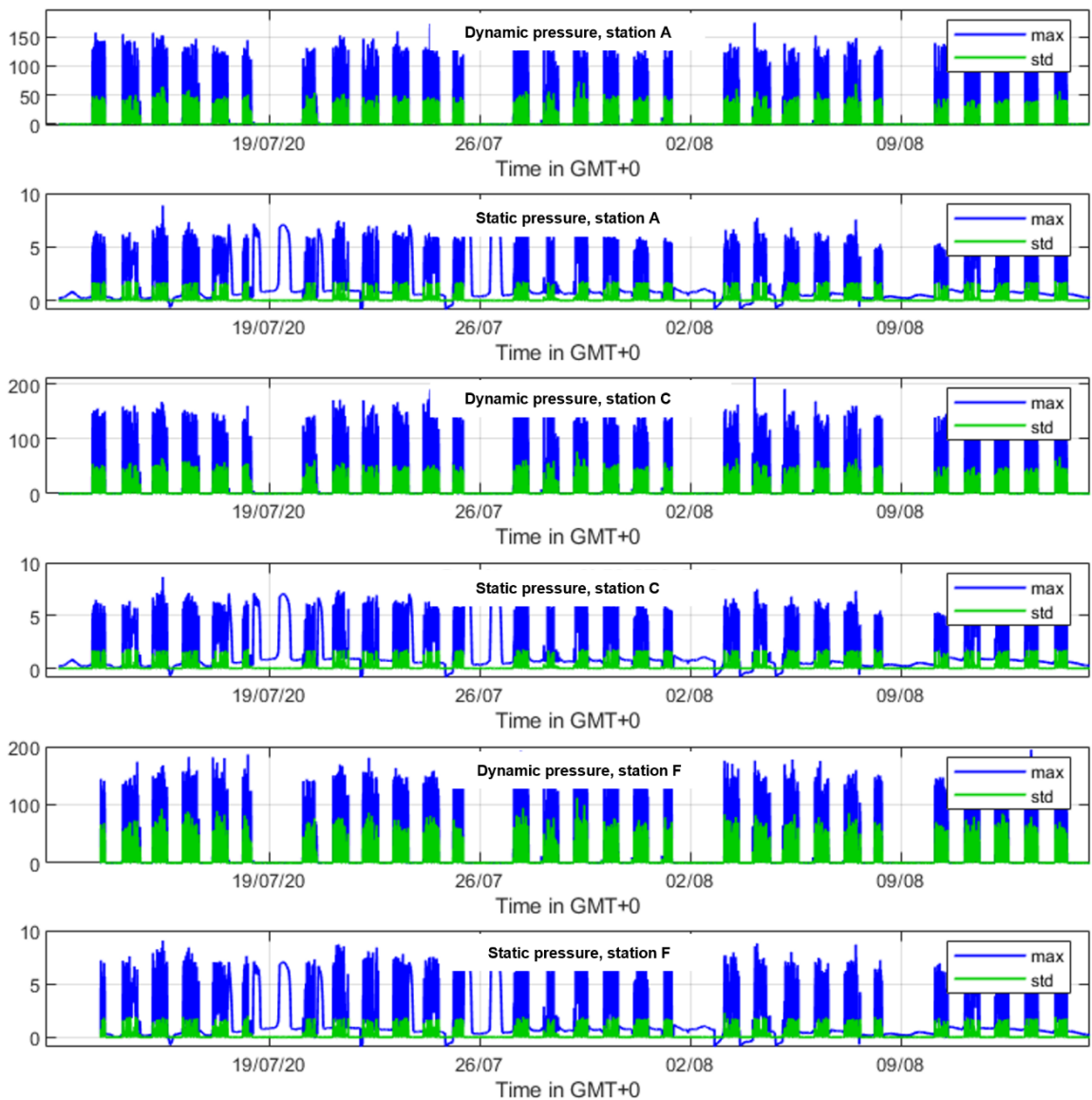

Fig. 3. Maximum and standard deviation of the static and dynamic pressures by hydrophones located at stations A, C and F.

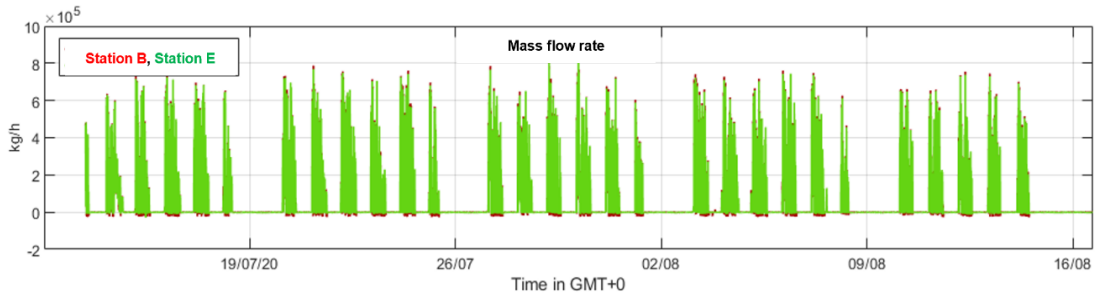

Fig. 4. Flow rate measured at the stations B and E.

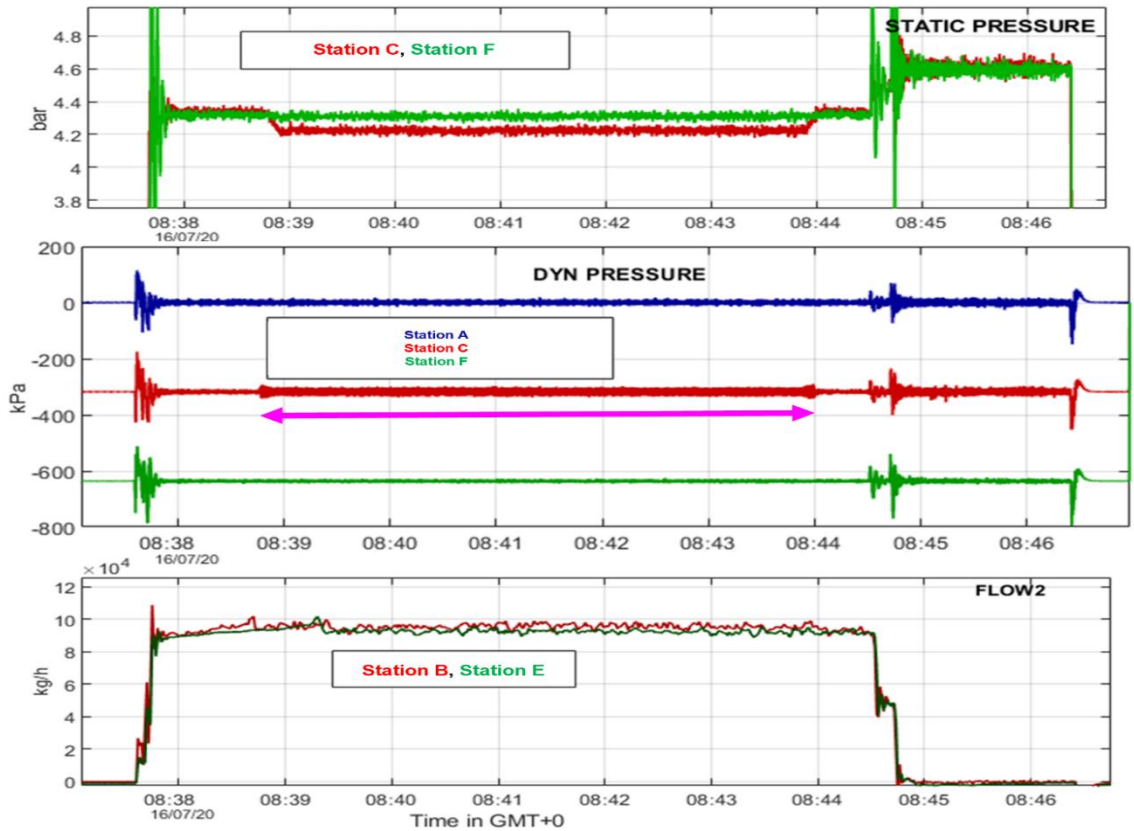

Fig. 5. From top to bottom, static pressure, dynamic pressure, and mass flow rate during a spill test. The magenta arrow indicates the duration of the controlled leak.

Successively, we have performed a clustering analysis of some pressure-related features recorded during the spill tests (e.g., fluid pressure, acceleration and flow rate, computed on specific time windows), with the objective of finding a parameter that is sensitive to the spill operation (e.g., whose values are different when the leak is in progress with respect to the non-leak case).

In addition, we have also compared the signals for the two spill locations (stations C and D). To ease the discussion, we report the most significant outcomes of all the tests performed:

- The acoustic noise produced by the exiting fluid at station D is visible at station D and F (maximum distance of approximately 50-60 m from the spill), but not at station A (250 m of distance from the tapping point).

- The acoustic noise produced by the exiting fluid at station C can only be observed very close to the leak point, not even at station A (63 m of distance from station C), probably due to the strong noise of the pumps that is superimposed to the signal of interest. It appears that the maximum detection distance of the acoustic noise is therefore below 60 m. During several tests, the absolute pressure is below 1 bar and the acoustic noise of the leak is not detectable.

- When pressure is stable and above 3-4 bar, there is a separation between the clusters related to different nozzles and the non-leaking operational status. It is then possible, in principle, to detect the presence of a spill. However, the vibroacoustic noise produced by the different tests apparently does not always exhibit a direct link with the differential pressure (in and out of pipe) and/or with the hole cross section, making it cumbersome to classify the leak. We think this reflects the different logistic scenarios of the tests.

- Acceleration signals offer a clearer separation between the clusters of different nozzles with respect to pressure signals. The localization of the spill position can be accurately obtained with a cross correlation procedure of the signal recorded on both sides of the leaking hole.

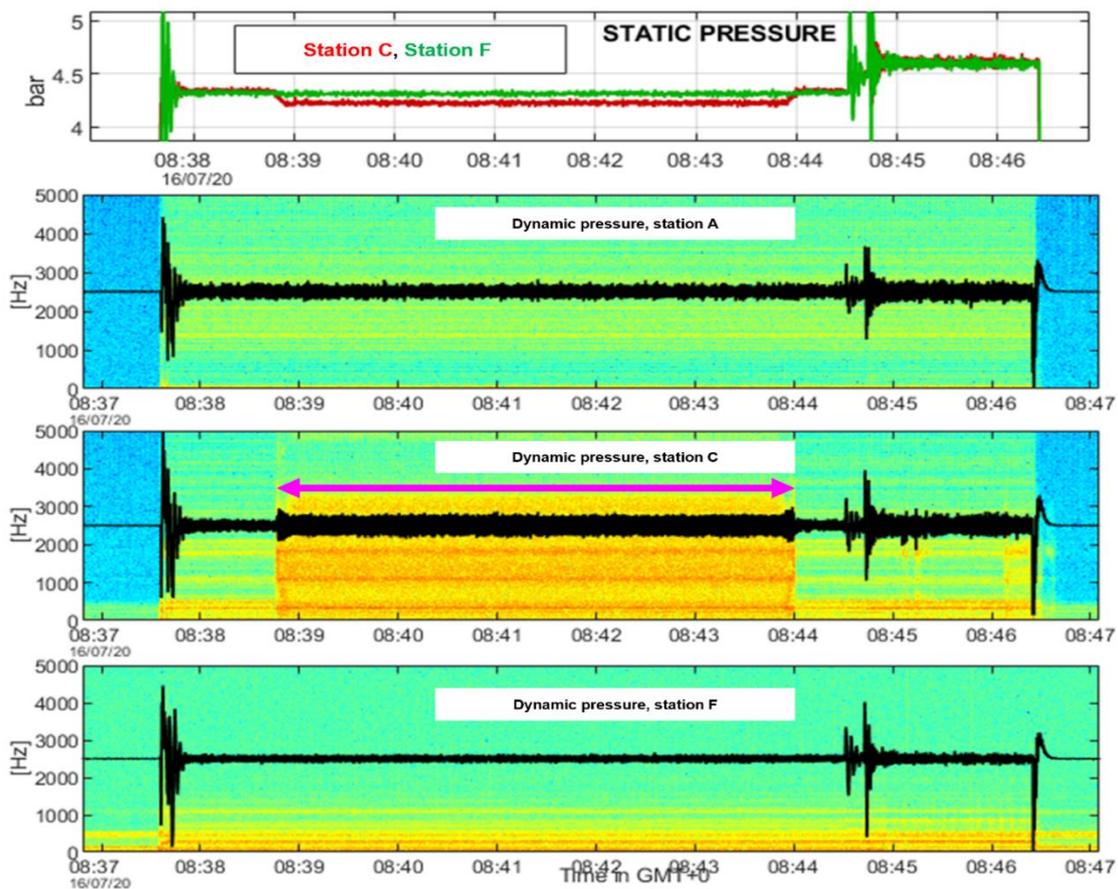

Fig. 6. Static pressure (top) and spectrogram of the dynamic pressure (colored images) with the overlaid time domain signal (black line). The spill noise (in the interval highlighted by the magenta arrow) is a high bandwidth signal, extending up to 4 kHz.

## 4. LEAKING HOLE DETECTION AND CLASSIFICATION

We explore here the possibility of detecting the fluid leaking from a hole in a pipe, by identifying the noise produced by the exiting flow. The analysis principle is within the acoustic noise (AN), leak detection systems (LDS) family.

When a hole is produced on a pipeline, there is fluid leakage through it. In case the fluid inside the pipe has a higher pressure with respect to the external one, we observe a turbulent exiting flow that is characterized by an irregular motion of particles and by velocity fluctuations at all points, interacting with the receiving environment. An acoustic noise is produced (also known as jet noise), equivalent to the radiation of dipoles/quadrupoles acoustic sources. The velocity of the jet $v_j$ can be derived starting from the following equation:

$$C_d = \frac{Q}{A_{or}} \sqrt{\frac{\rho}{2\Delta P}}, \qquad (1)$$

where $C_d$ is the orifice discharge coefficient, $Q$ is the flow rate, $A_{or}$ is the area of the orifice, $\rho$ is the density of the jet fluid and $\Delta P$ is the pressure drop. At low flow rates, $C_d$ is considered to be a function of the aspect ratio of the hole, the ratio between the diameter $d$ of the orifice and the diameter $D$ of the pipe, and of the Reynolds number of the orifice $R_e$. At high values of $R_e$, the effects of aspect ratio and Reynolds number become less relevant, as $C_d$ depends primarily on the diameter ratio. The jet velocity $v_j$ is then expressed as:

$$v_j = \sqrt{\frac{2\Delta P}{\rho}} = \frac{Q}{C_d A_{or}} \quad [\text{m/s}]. \qquad (2)$$

Based on the analysis and the limits presented in Section 3, we have then selected a subset of the tests, looking for an experimental relation between the leak induced vibroacoustic noise and the fluid flow parameters: more specifically, we have discarded some tests due to an absence of fluid flow (the system is off, absolute pressure is close to 0 bar) and also the spills performed at station C, since they were not visible at the closest station (e.g., A) due to the local noise of the operating pumps.

In addition, we have decided to classify the holes based only on their area and not on their shape. As a result of this operation, we have obtained three possible leak classes: small (area = 5.06 mm²), medium (area = 12.56 mm²) and large hole (area = 31.65 mm²).

Given the aforementioned considerations, we have observed that under similar external conditions (e.g. length, layout and filling status of purging pipe), the acoustic noise is proportional to the leaking mass, which in turn is proportional to the pressure gradient at the exiting point and to the hole cross section. Following Xu's approach [6], we propose a relation for the sound pressure level ($SPL$) of the type

$$SPL = \Delta p \cdot A^n \cdot k, \qquad (3)$$

where SPL is expressed in kPa, $\Delta p$ is the differential pressure (bar), $A$ is the hole area (mm²), $k$ is a scale factor (kPa/(bar·mm²)), and $n = (1.5, 2)$. The parameters $n$ and $k$ must be estimated experimentally.

The approach to detect and classify leaks consists in grouping the data (e.g., pressure, flow rate, etc.) into consecutive batches, having a predetermined temporal duration. On each of those time intervals, some statistical parameters are evaluated.

Every batch is then tagged with a specific color, according to its corresponding test scenario (small, medium, large hole or no leak). To increase the number of examples in the dataset, we have allowed a small overlap between consecutive time intervals. In some instances, the measurements corresponding to the same test have been averaged into a single, multidimensional variable.

The results of the detection and classification procedure are reported in Fig. 7, Fig. 8 and Fig. 9. The detection capability is satisfactory, as all the analyzed spills are detected. The accuracy in the estimation of the leaking hole area depends on the stationarity of the transportation system parameters during the spill action, e.g., the tests performed without fuel filling at the truck station provided the best results. Moreover, the bigger errors happen with the smallest hole, namely for the lowest SNR.

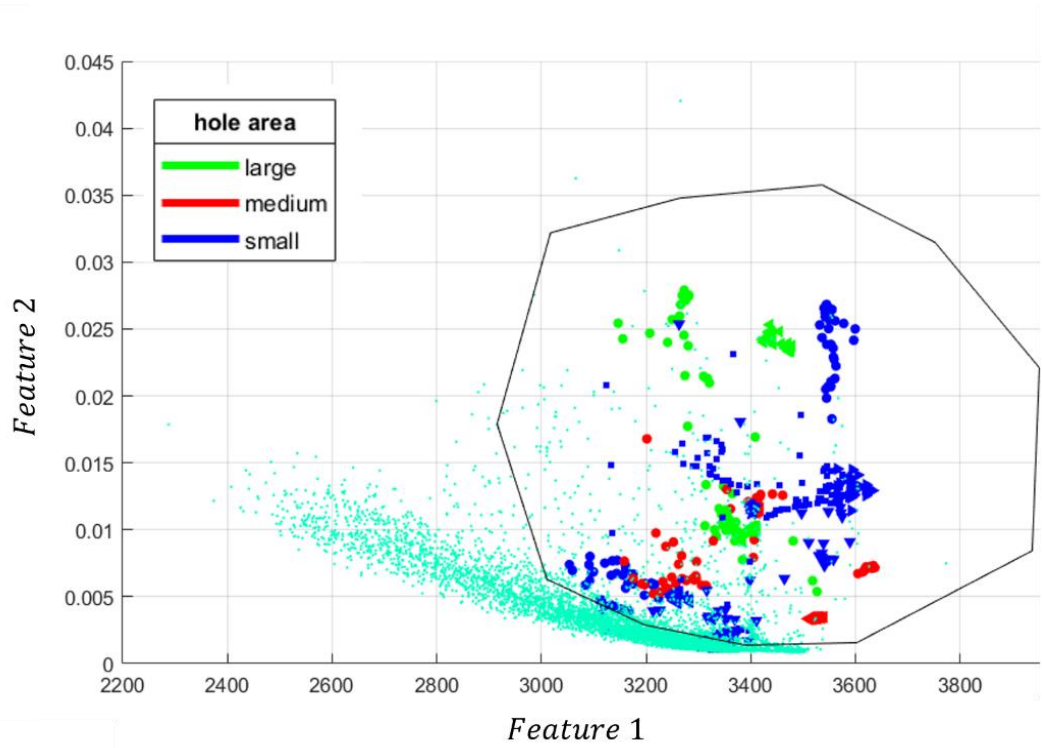

Fig. 7. Cross plot between two features and definition of a domain containing the spill cases (identified by the black polygon).

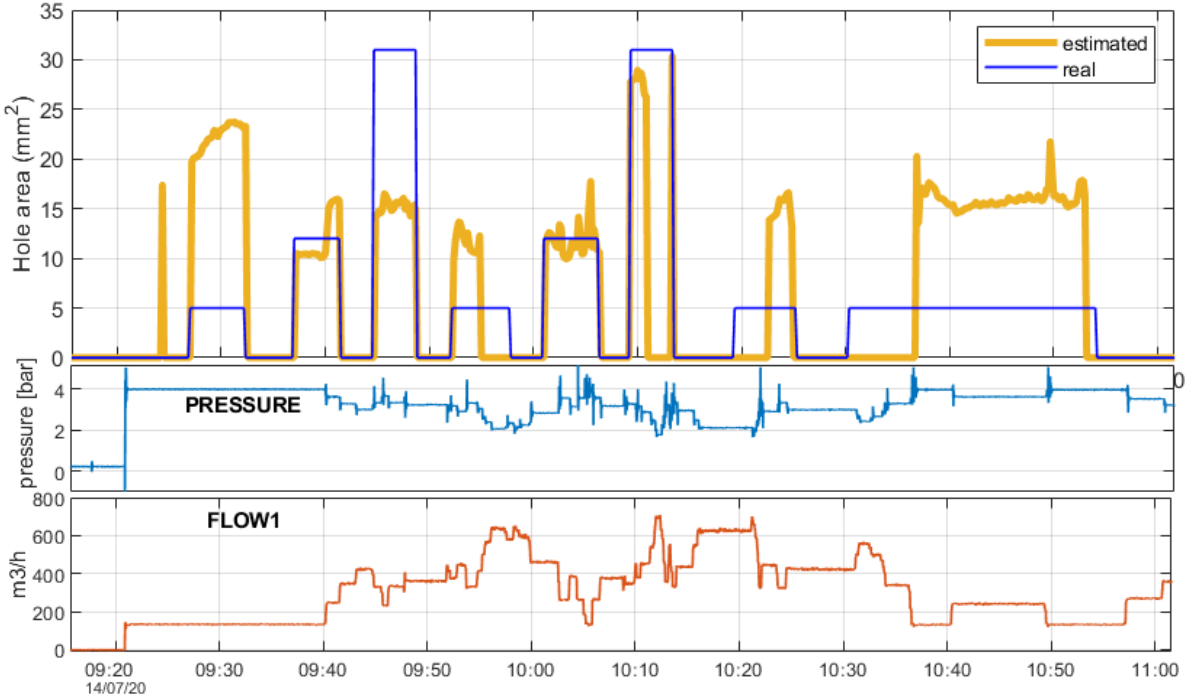

Fig. 8. Leaking hole detection and classification results (top) along with the corresponding flow and pressure conditions during the test.

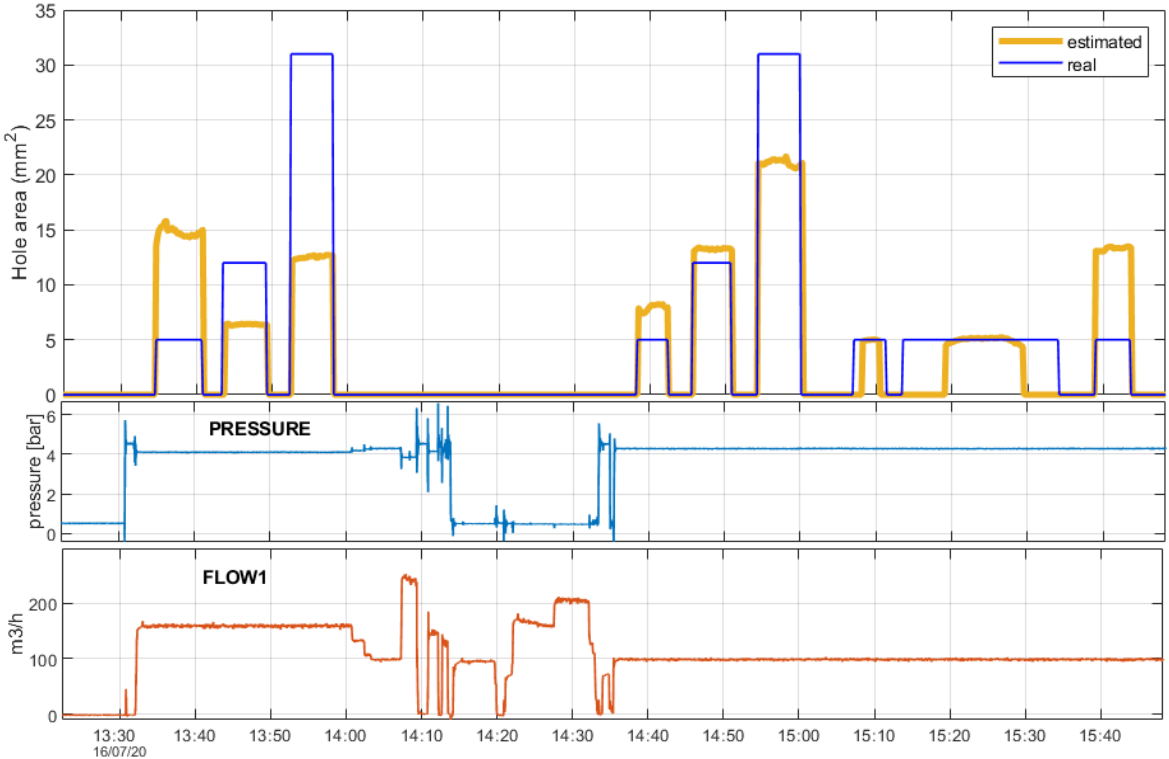

Fig. 9. Leaking hole detection and classification results (top) along with the corresponding flow and pressure conditions during the test.

## 5. CONCLUSION

We have presented a study on the detectability of a leaking hole in a pipeline, by exploiting the pressure transients generated by the escaping fluid (acoustic noise), and we have proposed a detection and classification procedure. The escaping fluid generates a medium/high frequency noise (above 500 Hz) that is detectable over the operational noise, even with active flow and for the smallest tested nozzle (having a cross section equal to 5 mm$^2$). The maximum detection distance is below 60 m when using pressure signals in a storage plant by connection piping system, due to the high spatial density of pumps and flow regulation devices. However, this last value can be increased by properly addressing the former issue, a cross correlation approach between the acoustic signal recorded on both sides of the spillage hole is sufficient and effective for its accurate localization.

The experimental data sets demonstrate that the hole shape cannot be reliably obtained by using the acoustic noise. However, there is a general direct link among the vibroacoustic noise measured as the standard deviation of the recorded signal, the in-out of pipe differential pressure, and the spillage hole area.

Experimental relations can be derived for the acoustic noise generated by a leaking fluid, to perform a hole area estimation. An equation that links the sound pressure level within the fluid with the area of the hole itself have been proposed.

All the analyzed spillages have been detected: the accuracy in the estimation of the leaking hole area depends on the stationarity of the transportation piping parameters during the spill action. Moreover, the bigger errors happen with the smallest hole, namely for the lowest SNR. Lastly, a deterministic approach for detecting and classifying a leak hole has been discussed: the results show that such a solution is viable and satisfactory.

### ACKNOWLEDGMENT

This research was mainly carried out in the framework of the R&D project founded by Eni S.p.A. The authors are grateful to Eni Logistic Department and SolAres JV teams for technical support during the field tests.